\documentclass[10pt, conference]{IEEEtran}
\IEEEoverridecommandlockouts
\usepackage{cite}
\usepackage{amsmath,amssymb,amsfonts}
\usepackage{algorithmic}
\usepackage{graphicx}
\usepackage{textcomp}
\usepackage{amsmath}
\usepackage{mathtools}
\usepackage{footnote}
\usepackage{placeins}
\usepackage{amsmath}
\usepackage{textcomp}
\usepackage{multirow}
\usepackage{placeins}
\usepackage{footnote}
\usepackage{multicol}
\usepackage{tabulary}
\usepackage{graphicx}

\usepackage{subcaption}
\usepackage{caption}
\usepackage{multicol}
\usepackage{footnote}
\usepackage{multirow}
\usepackage{multicol}
\usepackage{tabulary}

\usepackage{xcolor}
\def\BibTeX{{\rm B\kern-.05em{\sc i\kern-.025em b}\kern-.08em
    T\kern-.1667em\lower.7ex\hbox{E}\kern-.125emX}}
\begin{document}

\title{OCFormer: One-Class Transformer Network for Image Classification}

\author{\IEEEauthorblockN{Prerana Mukherjee \IEEEauthorrefmark{1},  Chandan Kumar Roy\IEEEauthorrefmark{2}, Swalpa Kumar Roy\IEEEauthorrefmark{3}}
\IEEEauthorblockA{\textit{\IEEEauthorrefmark{1}School of Engineering, Jawaharlal Nehru University, Delhi, India}}
\IEEEauthorblockA{\textit{\IEEEauthorrefmark{2}Department of Computer Science and Engineering, Indian Institute of Information Technology, Sri City, AP, India}}
\IEEEauthorblockA{\textit{\IEEEauthorrefmark{3}Jalpaiguri Government Engineering College, West Bengal, India}}}
%%

% \author{\IEEEauthorblockN{ Prerana Mukherji}
% \IEEEauthorblockA{\textit{Assistant Professor at IIIT Sricity} \\
% City, India \\
% \and
% \IEEEauthorblockN{ A N Vaishnavi }
% \IEEEauthorblockA{\textit{Undergraduate student at IIIT Sricity} \\
% Hyderabad, India \\
% vaish9699@gmail.com}
% \and
% \IEEEauthorblockN{ Gattineni Sai Sreenithya}
% \IEEEauthorblockA{\textit{Undergraduate student at IIIT Sricity} \\
% Hyderabad, India \\
% \and

% \IEEEauthorblockN{ Gullapalli Keerti}
% \IEEEauthorblockA{\textit{Undergraduate Student at IIIT Sricity} \\
% City, India \\
% \and
% \IEEEauthorblockN{A Sree Vidya}
% \IEEEauthorblockA{\textit{Undergraduate student at IIIT Sricity} \\
% Hyderabad, India \\
% \and
% \IEEEauthorblockN{ Deeksha Nayab}
% \IEEEauthorblockA{\textit{Undergraduate student at IIIT Sricity} \\
% Hyderabad, India \\

\maketitle

\begin{abstract}
We propose a novel deep learning framework based on Vision Transformers (ViT) for one-class classification. The core idea is to use zero-centered Gaussian noise as a pseudo-negative class for latent space representation and then train the network using the optimal loss function. In prior works, there have been tremendous efforts to learn a good representation using varieties of loss functions, which ensures both discriminative and compact properties. The proposed one-class Vision Transformer (OCFormer) is exhaustively experimented on CIFAR-10, CIFAR-100, Fashion-MNIST and CelebA eyeglasses datasets. Our method has shown significant improvements over competing CNN based one-class classifier approaches.

% In ever changing world the attempt is to address one of the significant issues that algorithmically made music had (and still has): the absence of worldwide rationality or structure. 
% In this paper, we will demonstrate our way to deal with producing old style music with rehashed melodic structures utilizing a Long Short Term Memory (LSTM) Neural Network with Attention.

% Attention is a mechanism combined  in the RNN enabling it to concentrate on specific pieces of the information arrangement while anticipating a specific piece of the yield grouping, empowering simpler learning and of higher caliber.

% We proposed the utilization of LSTMs with attention â€” which are evidently superior to vanilla-Recurrent neural network(RNNs) and LSTM for adapting longer fleeting conditions.Because of this experimentation, music has been one of the early utilizations of LSTMs.

%  In this way, with enough information and the right calculation, AI ought to have the option to make music that would sound human. This report plots different ways to deal with music creation through Neural Network models, and despite the fact that there were some blended outcomes by the model, it is clear that melodic thoughts can be gathered from these calculations in order to make another bit of music. 

\end{abstract}


\section{Introduction}
The primary objective for multi-class classification is to label a given object class from a predefined set of object categories. On the contrary, one-class classification attempts to model the samples from the positive or the normal class in order to classify new samples as not belonging to the set of target class. There is complete absence of negative class (as in case of binary classification) making the classification more complex. Such classification could be highly effective in cases where there is highly imbalanced or skewed dataset i.e. there are very few examples in the minority class or in cases where no supervised algorithm can learn effectively the coherent structure of the object classes. One-class classification variants include anomaly identification, outlier or novelty detection. 

Kernel density based generative methods \cite{breunig2000lof, latecki2007outlier} are highly popular for modeling anomalous data. They assign high density to train set while low density data are identified as outliers during testing. It suffers from the curse of dimensionality issue as it may assign high density to background as well and thus density estimation is not performed accurately \cite{morningstar2021density}. Discriminative methods including one-class SVM (OC-SVM) \cite{scholkopf1999support}, one-class CNN (OC-CNN) \cite{oza2018one} or support vector data descriptor (SVDD) \cite{tax2004support} utilize the non-linear kernels for outlier detection. The bottleneck in both the generative and discriminative methods is that they cannot learn the high dimensional data well and have poor input data representational ability. Thus, deep learning based one-class classifier methods \cite{ruff2018deep, chalapathy2018anomaly, oza2018one, ji2021one, lu2018deep} have been quite successful in learning the data representation. Although the naive learning of deep features often falls prey to “hypersphere collapse” \cite{ruff2018deep}.  These problems were circumvented by contrasting network architectures \cite{ruff2018deep}, autoencoder pretraining \cite{ruff2018deep, chalapathy2018anomaly}, surrogate multiclass classification on simulated outliers \cite{hendrycks2018deep, golan2018deep, hendrycks2019using, roy2019hybridsn, bergman2020classification, zhang2021one} or injecting noises \cite{oza2018one}.

Attention based models like transformers \cite{vaswani2017attention} quite popular in natural language processing (NLP) however of-late their usage in computer vision problems have also become manifold. The dominant approach is to pretrain on the large-corpus and fine-tune on the smaller dataset \cite{devlin2018bert}.  Inspired by transformer scaling success in NLP, Dosovitskiy \textit{et al.} \cite{dosovitskiy2020image} proposed a method where they applied transformer directly to images with fewest possible modifications and were able to beat the state of the art benchmark. In this work,  we propose a novel pipeline with vision transformer (ViT) based one-class classifier that maps the input image features to a latent space where they are concatenated with zero-centered Gaussian data and fed through the classifier. To the best of authors knowledge we are the first to explore ViT in one-class classification application. Exhaustive experiments on benchmark classification datasets validated that the proposed approach has superior performance over other competing one-class classifiers. 

% Here, in this paper we propose a vision transformer (ViT) based one-class classifier that maps the input image features to a latent space where they are concatenated with zero-centered gaussian data and passed through the classifier for further classification. Our method significantly gains accuracy boost when used with large pretrained transformers. We also show that dimensionality of latent space can significantly boost the performance of the classifier (in ablation study). The application of ViT in one-class classification is not explored yet. 

\section{Related Works}

Most of the anomaly detection, novelty detection tasks perform the one-class classification under the hood. The early works in this anomaly detection used a probabilistic approach to score the anomaly with low probability\cite{yamanishi2004line, eskin2000anomaly, xu2010robust} as compared to positive class data. 
% Generally, anomaly detection tasks are categorised into following categories: i) Representation based methods like sparse-coding \cite{turk1991eigenfaces} and PCA \cite{wold1987principal} map the input image to a  feature space and then reconstruct the original image from learned features and calculate the reconstruction error ii) methods \cite{knorr2000distance, gu2019statistical,hautamaki2004outlier, pang2018learning} follow the distance based learning approach, iii)  mixture models \cite{pang2018learning, kriegel2008angle}, iv) density based methods \cite{kim2012robust, manzoor2018xstream, breunig2000lof}, v) deep learning based auto-encoders \cite{chen2017outlier, you2017provable, sabokrou2016video, chalapathy2019deep}, vi) ensemble methods \cite{chen2017outlier, liu2008isolation} and vii) adversarial learning \cite{deecke2018image, schlegl2017unsupervised, perera2019ocgan, erfani2017shared}. 
Statistical methods like OCSVM \cite{scholkopf1999support} are closely related to SVM where the classifier finds the suitable hyperplane separating the data from different classes. Since in one-class classification, data from the negative class is absent, it is assumed that the negative class is the origin of the coordinate axis and the classifier finds a hyperplane that best separates the positive class from the origin. Another statistical method similar to OCSVM that uses kernel trick is SVDD \cite{tax2004support}, which tries to find the hypersphere that encloses positive data (inliers) completely. Similar to OCSVM \cite{scholkopf1999support}, there is another method one-class Minimax probability machine (OCMPM) \cite{lanckriet2003robust} which attempts to maximise the distance between origin and learned hyperplane by achieving a tighter lower bound of the training data. Another method Dual Slope Minimax Probability Machine (DS-OCMPM) \cite{perera2018dual} which is an extended version of OCMPM \cite{lanckriet2003robust} , considers two hyperplanes and availability of clusters. it finds the second hyperplane by projecting the data onto it such that data projected have largest possible variance. Generalised one-class Discriminative Subspace(GODS) \cite{wang2019gods} extends the OCSVM formulation into two hyperplanes. 
% Positive hyperplane is learnt such that most of the positive data lies on the positive side of it and negative hyperplane is learnt such that most of the negative data lie on the negative side of it.
Deep learning based methods learn the distribution using neural networks. There are discriminative methods that are inspired from OCSVM and SVDD in terms of loss functions. They use regularization techniques to make themselves more powerful. Methods like one-class CNN (OC-CNN) \cite{oza2018one} comprises of two subnetworks, one is feature extractor and another is classifier. The classifier is trained using Out-of-distribution (OOD) datasets.  Feature extractor could be any CNN model and classifier could be a MLP or OCSVM. Our work is mostly inspired from OC-CNN \cite{oza2018one}. 
% Deep SVDD (DSVDD)\cite{ruff2018deep} learns a deep representation with the objective of enclosing the embeddings of positively labeled data (inliers) with the smallest possible hypersphere. Holistic approach to one-class (HRN)\cite{hu2020hrn} trains a deep learning network with negative log likelihood loss with regularization applied on the learned features. 
Other deep learning based approaches are generative methods \cite{schlegl2017unsupervised, zaheer2020old, sabokrou2018adversarially} based on denoising-autoencoders, generative adversarial networks, auto-regressive models etc.  Self-supervision based methods have also gained popularity in the field of one-class classification. \cite{sohn2020learning} learns and evaluates  the deep feature representation in a self-supervised way. It first learns the self-supervised representation of one-class data and then builds a classifier on the learnt representation. Adapting Pretrained Features for Anomaly Detection (PANDA) \cite{reiss2020panda} introduces a method to combat catastrophic collapse(feature deteriorating) in pretrained transfer learning for one-class classification using early stopping and elastic regularization techniques. Inspired by \cite{oza2018one}, we have used a ViT instead of CNN that significantly outperforms many of the state of the art one-class classifiers by a large margin. 

\begin{figure*}[!ht]
    \centering
    \includegraphics[scale=0.5]{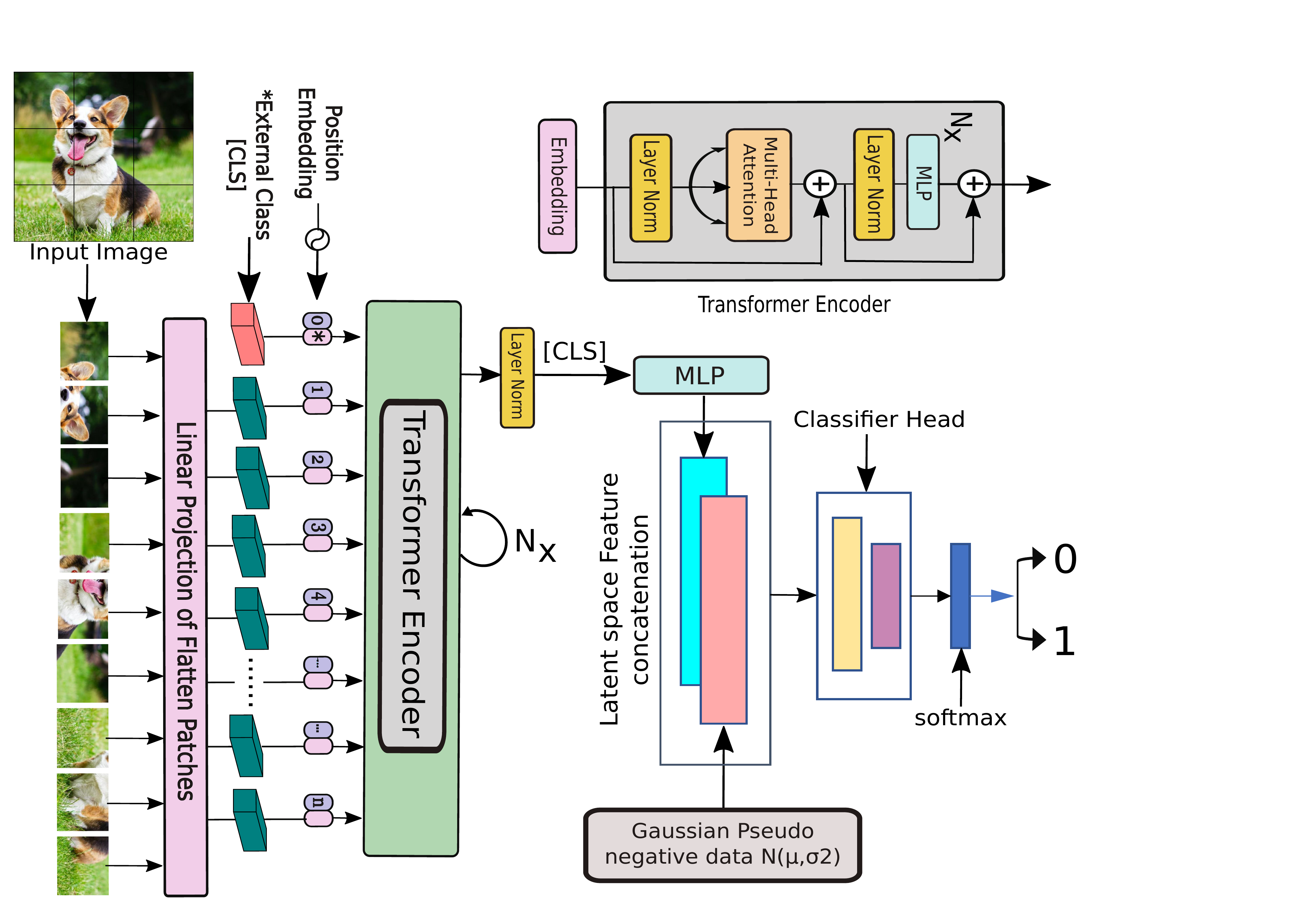}
    \caption{Graphical representation of the pre-trained one class transfomer. $\rm{\texttt{N}}_{\rm{\texttt{x}}}$ is 12 and 24 for ViT (base) and ViT (large), respectively.}
    \label{fig:prop}
\end{figure*}
%%%%%%%%%%%%%%%%%%%%%%%%%%%%%%%%%%%%%%%%%%%%%%%%%

\section{Proposed Methodology}
\label{section:methods}
We provide a detailed architecture of the proposed OCFormer, a one-class image classification approach in Fig.~\ref{fig:prop}. The network architecture consists of two key modules : i) feature extractor network and ii) classifier network. The feature extractor module basically maps the input target class images into a D dimensional feature space. Following the concatenation of these projected features with zero-centered Gaussian in the feature space, the features are fed into a classifier head which is typically a fully connected (FC) neural network. Classification scores are then assigned to each image by the classifier. The classification network returns either $0$ or $1$, with $0$ representing positive class and $1$ representing negative class. This notation assignment makes sense as we are essentially trying to recognise the abnormal one.

\subsection{Feature Extraction}
% Inspired congruent lines with \cite{oza2018one}, we conduct experiments with a different feature extractor that is Vision-Transformer (ViT) \cite{dosovitskiy2020image} instead of CNN. 
In ViT \cite{dosovitskiy2020image}, an image, $X$ is first divided into a sequence of patches, $X_{patches}$ using a 2D convolution operation having same kernel size and stride which ensures extracted features are non-overlapping in nature. Additionally, a classification token (\texttt{CLS}), $X_{cls}$ is appended to the patches. \texttt{CLS} is essential because its responsible for learning the abstract feature representation across the patches. To retain the positional information of the input patches with respect to the entire image, a positional embedding, $X_{pos}$ is added to all the patches including the \texttt{CLS} to obtain the input patch tokens, $X_{tokens}$. 
\begin{equation}
    \begin{aligned}
   X_{tokens} = [X_{cls} || X_{patches}] \oplus X_{pos}
   \end{aligned}
\end{equation}
These tokens are passed through sequential transformer encoder blocks. Transformer encoders are composite of blocks containing multiheaded self-attention (MSA) with a multi layer perceptron (MLP). In MLP, two multilayer perceptrons are used having an expanding ratio $r$ at the hidden layer, and one Gaussian Error Linear Unit (GELU ) activation is appended following the first linear layer. A layer normalization (LN) step precedes every block and residual shortcuts follow every block. 
\begin{equation}
\begin{aligned}
    X_{att} &= X_{tokens} \oplus MSA(LN(X_{tokens}))
    \\
    X' &= X_{att} \oplus MLP(LN(X_{att}))
\end{aligned}
\end{equation}
where $X'$ is the output of a transformer encoder block. After multiple transformer encoder blocks, the \texttt{CLS} token is sufficiently enriched with contextual information which can be used for the downstream tasks.

In this paper, we utilize the pre-trained ViT \cite{dosovitskiy2020image} network where 12 transformer encoder blocks are used in ViT (base) and 24 transformer encoder blocks are used in ViT (large). The embedding dimensions used are 768 and 1024 for ViT (base) and ViT (large) respectively. During the feature extraction phase, we freeze all layers excepts the classification head (i.e the last linear layer). The underlying assumption is that mapped features have D dimensions and are appended to the pseudo-negative data having Gaussian distribution $\mathcal{N}(\mu, \sigma^2,I)$, where $\mu$ and $\sigma^2$ correspond to the parameters of the Gaussian distribution, and $I$ corresponds to a $D\times{D}$ identity matrix. Here the distribution generates $D$ independent $1-D$ Gaussian with standard deviation.

\subsection{Classification Network}

As we append the pseudo-negative data, the classifier network observes the data in twice the batch size of the initial. So, if the initial input has a batch size of $B$, then the classifier will get the input features having size of $2B$. An FC layer and a softmax regression layer are used together in a sequence which act as the classifier layer. The FC layer has the same dimensionality as the input feature. The output of the softmax layer is two. 

\subsection{Loss Function}
To train the network, we have utilized binary cross entropy loss given as, $L_c=-\frac{1}{2K}\sum_{j=1}^{2K}(y log(p)+(1-y)log(1-p))$
% \begin{equation}
%     L_c=-\frac{1}{2K}\sum_{j=1}^{2K}(y log(p)+(1-y)log(1-p))
% \end{equation}
where $y \in (0, 1)$ and $y$ = 0 indicates the classifier input comes from the feature extractor, and $y$ = 0 denotes that the classifier input is sampled from Gaussian distribution. $p$ denotes the softmax probability with $y$ = 1.
We utilized Adam optimizer \cite{kingma2014adam} with standard learning rate and the weight decay is set as 1e-4. The batch size for the input images is set to 64. $\mu$ is set to 0 and $\sigma^2$ is set to 0.01. In order to stabilisethe training process, we utilized instance normalization \cite{ulyanov2016instance}. We fixed the latent dimension D as 1000, as it was found to be more promising. 

\section{Experiments}

\subsection{Datasets} 
In congruent lines with \cite{sohn2020learning}, we have evaluated our proposed approach on four one-class classification benchmark datasets including CIFAR-10, CIFAR-100 \cite{krizhevsky2009learning} and Fashion-MNIST \cite{xiao2017fashion} and CelebA Eyeglasses \cite{liu2015deep} dataset. Here in order to detect the abnormal class, we have used a one-vs-all approach. The anomaly class was one-class from all the classes and the normal class was taken randomly from the remaining classes equal to the normal class number. Random selection of training data points slightly introduces class imbalance problem that makes this more challenging. \textbf{CIFAR10} consists of 60,000 colour images with 32x32 dimensionality across 10 classes (6000 images per class). 50000 are training images and rest are used as testing images. Using leave one out approach, 10 different classes of anomaly and normal classes were constructed. For each combination, there were 5k training examples and 1k testing examples per class. \textbf{CIFAR100} consists of 60,000 colour images with 32x32 dimensionality (600 images per class). 500 are used as training images and 100 testing images per class. Using leave one out approach, 20 different classes of anomaly and normal classes were constructed. \textbf{Fashion-MNIST} consists of 60,000 gray scale images with 28x28 dimensionality across  10 different classes, along with a  set of 10000 testing images. For this dataset too, 10 different combinations of training and testing sets were considered using the same leave one out method. For each training sample there were 6k training samples and 1k testing samples for each class. \textbf{CelebA eyeglasses} consists of 202599 images of celebrities out of which 13193 images are of celebrities with eyeglasses that are treated as outliers. A total of  17163 images without eyeglasses are included in the training dataset and 26386 images in the test data having equal proportion of both the outliers and the inliers data.

\subsection{Baseline methods} 
We utilized the following baselines to compare with our model. HRN \cite{hu2020hrn} that uses holistic regularization applied for one class learning. DSVDD\cite{ruff2018deep} a deep learning based kernel method which learns a deep representation in order to enclose the embeddings of positively labeled data (inliers) contained within the smallest possible hypersphere. To discriminate between transformations applied to images, Golan \textit{et al.} \cite{golan2018deep} trained a multi-class model. Bargman and Hoshen \cite{bergman2020classification} unifies the transformation and classification based methods and extends them for a broad range of applications on non-image data. Hendrycks \textit{et al.} \cite{hendrycks2019using} proposed a self supervised technique. Huang \textit{et al.} \cite{huang2019inverse} proposed a transformation based method in which a certain information from the normal image is erased based on human priors and expects an inverse-transform model  to predict the missing information. HRN \cite{hu2020hrn} adds an additional regularization parameter to negative log likelihood loss. PANDA \cite{reiss2020panda} is a transfer-learning approach that deals with some of the drawbacks of supervised-learning methods using early stopping and elastic regularization by continual learning. MSCLAD \cite{reiss2021mean} utilized the catastrophic failures of other anomaly detection methods by using a new loss function that overcomes failure mode of both contrastive and center loss.

\subsection{Training details and hyperparameters selection}

OCFormer utilizes a pre-trained feature extractor (ViT) to learn the latent mapping of normal images. latent features are concatenated with pseudo negative data that is zero centered gaussian noise. It uses a MLP layer as a classifier (only 1 layer in our case). It uses Base-ViT (86M parameters)  as standard feature extractor but we extend it further to larger ViT (307M parameters). We use Adam optimizer with weight decay fixed as 1e-4. For each normal class of CIFAR10, CIFAR100 and Fashion-MNIST, we train it for over 15 epochs (where for CIFAR10 and CIFAR100, each epoch consists of 71 uniform steps and for the Fashion-MNIST, each epoch consists of 85 steps) and for the CelebA eyeglass dataset, we train it for 10 epochs (each epoch consists of 2616 steps) with batch size of 64 as beyond 10 epochs, model starts overfitting and performance decreases significantly. We use input image size of 224$\times$224 to match the input size of pretrained ViT. We fix the latent dimension (D) representation to 1000. 
\vspace{-0.5cm}
\subsection{Quantitative and Qualitative Results} 
We report the results on four image datasets namely CIFAR-10, CIFAR-100, F-MNIST and CelebA eyeglasses.  We report area under curve (AUC-ROC which is a benchmark evaluation metric for one-class classification) for each of the dataset averaged over all the classes.  Each column in table \ref{tab:tab1} represents the mean AUC-ROC (\%) with the standard deviation (std) calculated over 5 runs. In table \ref{tab:tab1} our proposed method OCFormer outperforms all the existing state-of-the-art one-class classification approaches.  PANDA \cite{reiss2020panda} and MSCALD \cite{reiss2021mean} are competitive to our method. However when ViT-large with more parameters is used, it outperforms existing methods. This generalization of ViT can be explained as the number of parameters gets increased, the model is open to learn all the complex functions though there is a slight chance of overfitting but it can be avoided by using regularization techniques.  HRN \cite{hu2020hrn} performs similar to our model in case of f-MNIST but it  fails when it comes to CIFAR10 and CIFAR100. Other methods like Deep-SVDD \cite{ruff2018deep} performs reasonably poorly on the CIFAR10 dataset. Bargman and Hoshen \cite{bergman2020classification} is quite close in performance with  OCFormer on f-MNIST dataset but it is outperformed by OCFormer in CIFAR-10.

% Last row (in bold) is our result that outperforms all the existing state-of-the-art model results for one-class classification. More can be found in section 3 (comparison to previous work).
% \vspace{-0.8cm}
\begin{table}[]
\caption{ Performance evaluation (ROC-AUC\%-mean and std are calculated over 5 runs) with test time data augmentation and compare them with various one-class classification benchmark methods.
}
\footnotesize
\label{tab:tab1}
\begin{tabular}{lccc}
\hline
\textbf{Method}                                                                           & \multicolumn{1}{l}{\textbf{CIFAR-10}}                                                        & \multicolumn{1}{l}{\textbf{CIFAR-100}}                                & \multicolumn{1}{l}{\textbf{f-MNIST}}                                    \\ \hline
Deep-SVDD\cite{ruff2018deep}                                                                         & 64.8                                                                                         & -                                                                     & -                                                                       \\ \hline
Golan and EI-Yaniv \cite{golan2018deep}                                                                & 86.0                                                                                         & 78.7                                                                  & 93.5                                                                    \\ \hline
Bargman and Hoshen \cite{bergman2020classification}                                                                 & 88.2                                                                                         & -                                                                     & 94.1                                                                    \\ \hline
Hendrycks et. al.\cite{hendrycks2019using}                                                                  & 90.1                                                                                         & -                                                                     & -                                                                       \\ \hline
Huang et. al.\cite{huang2019inverse}                                                                     & 86.6                                                                                         & 78.8                                                                  & 93.9                                                                    \\ \hline
HRN\cite{hu2020hrn}                                                                                & 71.32                                                                                        & \multicolumn{1}{l}{}                                                  & 92.84                                                                   \\ \hline
PANDA\cite{reiss2020panda}                                                                              & 96.2                                                                                         & 94.1                                                                  & -                                                                       \\ \hline
MSCLAD\cite{reiss2021mean}                                                                           & 97.5                                                                                         & 96.5                                                                  & -                                                                       \\ \hline
\textbf{\begin{tabular}[c]{@{}l@{}}OCFormer (ViT-base)\\ OCFormer (ViT-large)\end{tabular}} & \multicolumn{1}{l}{\begin{tabular}[c]{@{}l@{}}96.77±0.02\\ \textbf{98.72±0.40}\end{tabular}} & \begin{tabular}[c]{@{}c@{}}95.08±0.1\\ \textbf{96.7±0.3}\end{tabular} & \begin{tabular}[c]{@{}c@{}}92.718±0.3\\ \textbf{94.34±0.3}\end{tabular} \\ \hline
\end{tabular}
\end{table}

\begin{table}[]
\caption{ Comparative result (mean and std are calculated over 5 runs) with \cite{sohn2020learning}. 
}
\tiny
\label{tab:tab2}
\centering
\begin{tabular}{cccccc}
\hline
\textbf{Method}                                                                                & \textbf{Classifier}                                                                    & \textbf{CIFAR-10}                                                                                                            & \textbf{CIFAR-100}                                                                                                           & \textbf{f-MNIST}                                                                                                        & \textbf{CelebA}                                                                     \\ \hline
resnet50 \cite{sohn2020learning}                                                                                      & \begin{tabular}[c]{@{}c@{}}OC-SVM\\ KDE\end{tabular}                                   & \begin{tabular}[c]{@{}c@{}}80.0\\ 80.0\end{tabular}                                                                          & \begin{tabular}[c]{@{}c@{}}83.7\\ 83.7\end{tabular}                                                                          & \begin{tabular}[c]{@{}c@{}}91.8\\ 90.5\end{tabular}                                                                     & \begin{tabular}[c]{@{}c@{}}81.4\\ 82.4\end{tabular}                             \\ \hline
RotNet \cite{sohn2020learning}                                                                                & \begin{tabular}[c]{@{}c@{}}Rotation classifier\\ KDE\end{tabular}                      & \begin{tabular}[c]{@{}c@{}}86.8±0.4\\ 89.3±0.3\end{tabular}                                                                  & \begin{tabular}[c]{@{}c@{}}80.3±0.5\\ 81.9±0.5\end{tabular}                                                                  & \begin{tabular}[c]{@{}c@{}}87.4±1.7\\ 94.6±0.3\end{tabular}                                                             & \begin{tabular}[c]{@{}c@{}}51.4±3.9\\ 77.4±1.0\end{tabular}                             \\ \hline
Denoising  \cite{sohn2020learning}                                                                                    & \begin{tabular}[c]{@{}c@{}}OC-SVM\\ KDE\end{tabular}                                   & \begin{tabular}[c]{@{}c@{}}83.4±1.0\\ 83.5±1.0\end{tabular}                                                                  & \begin{tabular}[c]{@{}c@{}}75.2±1.0\\ 75.2±1.0\end{tabular}                                                                  & \begin{tabular}[c]{@{}c@{}}93.9±0.4\\ 93.7±0.4\end{tabular}                                                             & \begin{tabular}[c]{@{}c@{}}66.8±0.9\\ 67.0±0.7\end{tabular}                                                                              \\ \hline
Rotation prediction \cite{sohn2020learning}                                                                           & \begin{tabular}[c]{@{}c@{}}OC-SVM\\ KDE\end{tabular}                                   & \begin{tabular}[c]{@{}c@{}}90.8±0.3\\ 91.3±0.3\end{tabular}                                                                  & \begin{tabular}[c]{@{}c@{}}82.8±0.6\\ 84.1±0.6\end{tabular}                                                                  & \begin{tabular}[c]{@{}c@{}}94.6±0.3\\ 95.8±0.3\end{tabular}                                                             & \begin{tabular}[c]{@{}c@{}}65.8±0.9\\ 69.5±1.7\end{tabular}                                                                               \\ \hline
Contrastive \cite{sohn2020learning}                                                                           & \begin{tabular}[c]{@{}c@{}}OC-SVM\\ KDE\end{tabular}                                   & \begin{tabular}[c]{@{}c@{}}89.0±0.7\\ 89.0±0.7\end{tabular}                                                                  & \begin{tabular}[c]{@{}c@{}}82.4±0.8\\ 82.4±0.8\end{tabular}                                                                  & \begin{tabular}[c]{@{}c@{}}93.9±0.3\\ 93.6±0.6\end{tabular}                                                             & \begin{tabular}[c]{@{}c@{}}83.5±2.4\\ 84.6±2.5\end{tabular}                             \\ \hline
Contrastive(DA) \cite{sohn2020learning}                                                                        & \begin{tabular}[c]{@{}c@{}}OC-SVM\\ KDE\end{tabular}                                   & \begin{tabular}[c]{@{}c@{}}92.5±0.6\\ 92.4±0.7\end{tabular}                                                                  & \begin{tabular}[c]{@{}c@{}}86.5±0.7\\ 86.5±0.7\end{tabular}                                                                  & \begin{tabular}[c]{@{}c@{}}\textbf{94.8±0.3}\\ \textbf{94.5±0.4}\end{tabular}                                                             & \begin{tabular}[c]{@{}c@{}}84.5±1.1\\ 85.6±0.5\end{tabular}                             \\ \hline
\textbf{\begin{tabular}[c]{@{}c@{}}OCFormer(ViT-base)\\ \\ \\ \\ OCFormer(ViT-large)\end{tabular}} & \textbf{\begin{tabular}[c]{@{}c@{}}MLP\\ SVM\\  KDE\\ Twin-SVM \\ \\ MLP\\ SVM\\ KDE\\Twin-SVM\end{tabular}} & \begin{tabular}[c]{@{}c@{}}\textbf{96.77±0.02}\\ \textbf{97.06±0.3}\\ 80.4±0.4\\  82.36±0.3\\ \\ \textbf{98.72±0.4}\\ \textbf{97.99±0.7}\\ 84.36±0.3\\ 85.24±0.6
\end{tabular} & \begin{tabular}[c]{@{}c@{}}\textbf{95.08±0.1}\\ \textbf{94.31±0.5}\\     73.01±0.3\\75.48±0.5
\\ \\ \textbf{96.7±0.3}\\ \textbf{95.65±0.6}\\ 82.24±0.3\\ 79.60±0.5\end{tabular} & \begin{tabular}[c]{@{}c@{}}92.718±0.3\\ 88.53±0.5\\ 80.14±0.2\\ 75.96±0.7
\\ \\ 94.34±0.3\\ 93.31±0.6\\ 77±0.2\\80.91±0.3
\end{tabular} & \begin{tabular}[c]{@{}c@{}}80±1.2 \\\textbf{94.86}\\ \textbf{93.30}\\70.54±.6\\ \\ \textbf{96.58}\\ \textbf{95.65}\\72±1.0\\74.34±0.8
 \end{tabular}\\ \hline
\end{tabular}
\end{table}

\begin{figure*}[thpb]
      \centering
      \fbox{
      \includegraphics[scale=1]{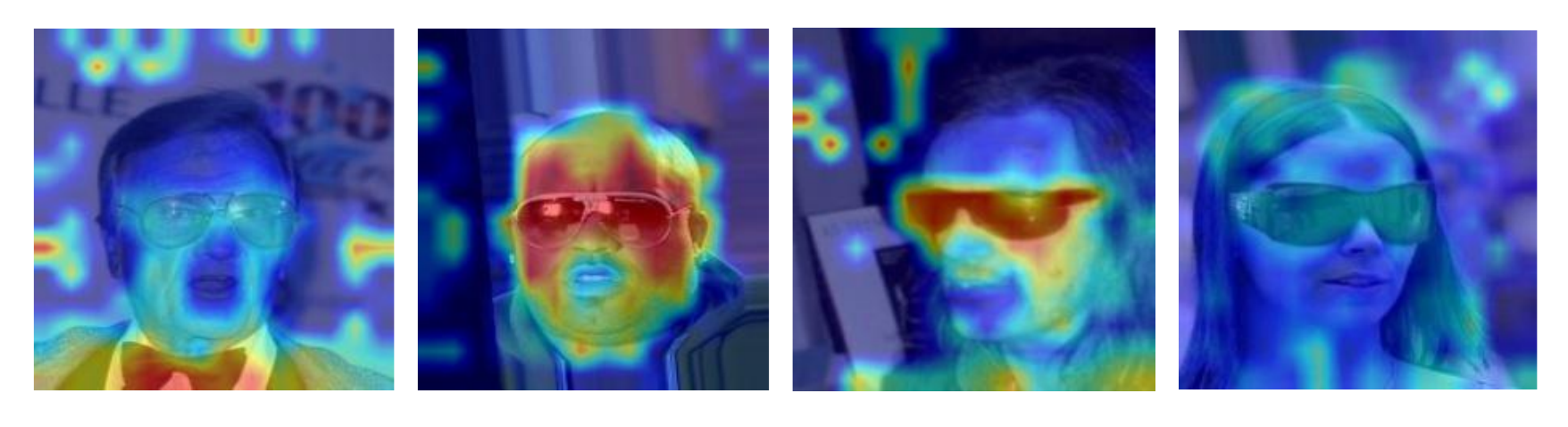}}
      \caption{GradCAM activation map \cite{selvaraju2017grad} for few sample images from CelebA eyeglasses dataset}
      \label{fig:qualitative}
\end{figure*}

\begin{figure}[thpb]
      \centering
      \fbox{
      \includegraphics[scale=0.3]{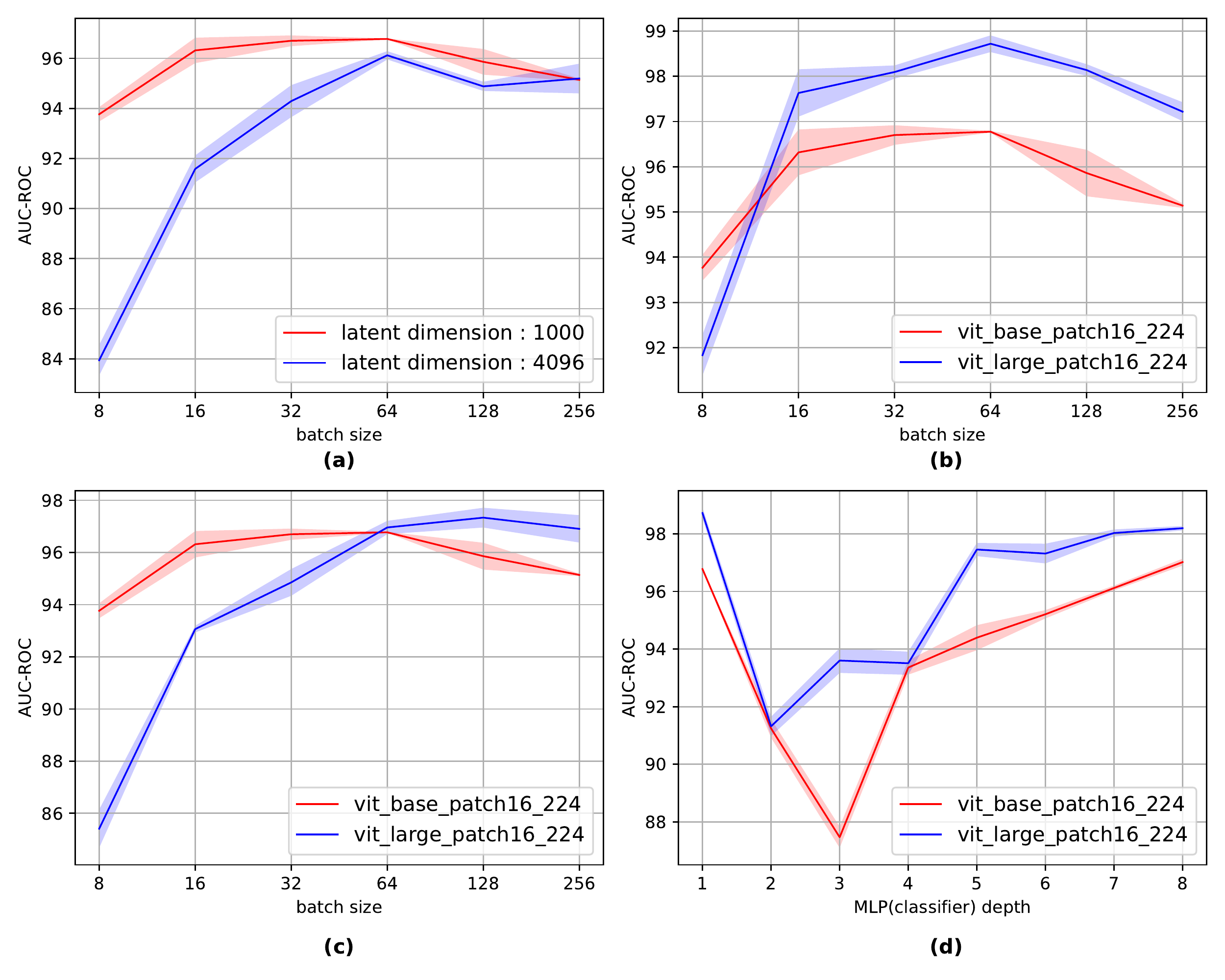}}
      \caption{Effect of batch size, latent dimension, classifier depth and feature extractor on AUC using Cifar10 dataset. Ablation Studies with a) Area Under the Curve plot (AUC) with various batch sizes and latent dimensions,  b) with varying  ViT (D=1000)         c) with varying ViT (D=4096) and d) with varying MLP classifier depth. 
}
      \label{fig:graph}
\end{figure}

We compare our model performance with the results in \cite{sohn2020learning} in table \ref{tab:tab2}.  The mean of each dataset is weighted by the number of classes. It can be observed that OCFormer (ViT-base and ViT-large) consistently outperforms all other methods in terms of mean AUC\% in CIFAR10 and CIFAR100 using MLP and SVM as classifiers by a relative margin of 6.7\%. For CelebA eyeglasses dataset OCFormer achieves best performance with other compared methods with MLP (in both ViT-base and ViT-large), SVM (in both ViT-base and ViT-large) and KDE (in ViT-base). In Fig. \ref{fig:qualitative}, we provide visual explanations using GradCam activations which shows that the attention is on the eyeglasses worn by people (which is the abnormal class).    
% Resnet50 pre-trained on ImageNet is used as a feature extractor, achieves 84.2 and 84.15 mean AUC over three datasets using OC-SVM and KDE as classifiers respectively. Other self-supervised approaches of \cite{sohn2020learning} performs better than the resnet50, RotNet \cite{golan2018deep} achieves nearly 76.4 and  mean AUC while Denoising, Rotation Prediction, Contrastive learning without and with distributional augmentation result mean AUC of 84.83, 84.16, 89.4, 88.40 and 91.26 respectively.
% /In our case, representation learnt using fine-tuning  the pre-trained ViT base outputs mean AUC of 94.86 using MLP classifier and 93.30 using SVM classifier. With larger ViT (307M parameters) mean AUC increases upto 96.58 using MLP classifier and 95.65 using SVM classifier.

\subsubsection{ABLATION STUDY AND ANALYSIS}
We analyse the behaviour  and performance of one-class classification and in Fig. \ref{fig:graph} we provide the ablation results on  CIFAR10 dataset.

\textbf{Batch size.} We tried to explore the robustness of our method using varying batch sizes in Fig. \ref{fig:graph} (a), latent dimensions and feature extractor ViT with increasing parameters. We train the model with various batch sizes {$2^3, \ldots, 2^8$}. We carefully studied the behaviour of the model with respect to batch size and the model complexity. Performance of a model is highly affected by the batch size. Smaller batch sizes may not generalize well and larger batch size does not guarantee a better representational learning of the training data. 

\textbf{Base feature extractor.} with pre-trained ViT-base (86M parameters) shows the mean AUC of 91.16 over all the four datasets with MLP head as classifier while the performance improves when we use pre-trained ViT-large (307M parameters) as a feature extractor with a similar set of experiments, the mean AUC comes out to be 96.58. This difference in  the performance could be explained as the number of parameters gets increased, model becomes flexible enough to map any function but the chance of overfitting increases. But with standard regularization techniques overfitting can be avoided.

\textbf{Latent dimension augmentation.}  With latent dimension as 4096, it performs poorer than latent dimension as 1000. It can be seen clearly in Fig. \ref{fig:graph}(a), the red colored line represents D as 1000 and blue line represents D as 4096. Irrespective of  the batch size being used, the red colored line always outperforms the blue curve. In Fig. \ref{fig:graph}(b) D as 1000 achieves its high performance as high as 98.72\% AUC-ROC on CIFAR-10 while in Fig. \ref{fig:graph}(c) D as 4096 is able to achieve its high performance AUC-ROC as 97\% even on larger base feature extractor.

\textbf{MLP Depth.} It can be observed in Fig. \ref{fig:graph}(d) that the performance initially drops with the increase in MLP classifier's for both the feature extractors. But as we go deeper in the classifier network, model's performance starts improving.

% \begin{figure}[h!]
%     \centering
%     \includegraphics[scale=0.2]{mnist-results.png}
%     \caption{Figure illustrates some adversarial example generated from our attack on MNIST given the original image. The label above figure gives the actual prediction of classifier given input image, and label below gives the prediction of humans followed by '->' with classifier prediction. }
%     \label{fig:mnist-results}
% \end{figure}
% \vspace{-0.2cm}
\section{Conclusion}
In this letter, we present a simple yet powerful two stage transformer based one-class classifier. The idea of latent dimension augmentation of features for abnormal classes opens avenues for traditional classifiers such as SVM, KDE into the domain of one-class classification. The proposed model outperformed compared one-class classifier approaches conveniently. A visual representation of the class activation maps is also included to explain the proposed model’s performance.

\bibliographystyle{IEEEtran}
\bibliography{refs}

\end{document}